\documentclass[10pt,twocolumn,letterpaper,table,dvipsnames]{article}
\pdfoutput=1 

\usepackage[utf8]{inputenc}
\usepackage[T1]{fontenc}
\usepackage{textcomp}

\usepackage{3dv}
\usepackage{times}
\usepackage{epsfig}
\usepackage{graphicx}
\usepackage{amsmath}
\usepackage{amssymb}

\usepackage{ifthen}
\newboolean{includeSuppMat}
\setboolean{includeSuppMat}{true} 
\newcommand{\ifIncludeSuppMat}[2]{\ifthenelse{\boolean{includeSuppMat}}{#1}{#2}}
\ifIncludeSuppMat{
}{
\usepackage{xr}
\externaldocument[supp-]{VideoDepthFromDefocus-supp}
}

\usepackage{microtype}
\usepackage{enumitem}
\usepackage{xspace}
\usepackage{wrapfig}
\usepackage{sidecap}

\usepackage[breaklinks=true,colorlinks,bookmarks=false]{hyperref}

\usepackage[numbers,sort]{natbib}
\usepackage[capitalise]{cleveref}
\Crefname{figure}{Figure}{Figures}
\crefformat{equation}{Equation~#2#1#3}
\crefrangeformat{equation}{Equations~#3#1#4 to~#5#2#6}
\crefmultiformat{equation}{Equations~#2#1#3}{ and~#2#1#3}{, #2#1#3}{ and~#2#1#3}

\DeclareMathOperator*{\argmin}{arg\,min}

\newcommand{\abs}[1]{\left\lvert#1\right\rvert}
\newcommand{\norm}[1]{\left\lVert#1\right\rVert}

\definecolor{orange}{rgb}{1.0,0.4,0}
\definecolor{skyblue}{rgb}{0.2,0.6,0.9}
\definecolor{darkgreen}{rgb}{0,.5,0}
\definecolor{placeholder}{rgb}{0.6,0.8,0.95}
\newcommand{\footurl}[2]{\href{#1}{#2}\footnote{\url{#1}}}

\newcommand{\inlineheading}[1]{\vspace{0.5em}\noindent\textbf{#1\hspace{0.5em}}}

\threedvfinalcopy 

\ifthreedvfinal\fi
\begin{document}

\title{Video Depth-From-Defocus}

\author{%
	Hyeongwoo Kim $^\text{1}$\qquad
	Christian Richardt $^\text{1, 2, 3}$\qquad
	Christian Theobalt $^\text{1}$%
	\\[1em]
	$^\text{1}$ Max Planck Institute for Informatics\quad%
	$^\text{2}$ Intel Visual Computing Institute\quad%
	$^\text{3}$ University of Bath%
}

\hypersetup{
	pdftitle={Video Depth-From-Defocus},
	pdfauthor={Hyeongwoo Kim, Christian Richardt, Christian Theobalt}
}

\maketitle

\begin{abstract}
Many compelling video post-processing effects, in particular aesthetic focus editing and refocusing effects, are feasible if per-frame depth information is available. 
Existing computational methods to capture RGB and depth either purposefully modify the optics (coded aperture, light-field imaging), or employ active RGB-D cameras.  
Since these methods are less practical for users with normal cameras, we present an algorithm to capture all-in-focus RGB-D video of dynamic scenes with an unmodified commodity video camera.
Our algorithm turns the often unwanted defocus blur into a valuable signal.
The input to our method is a video in which the focus plane is continuously moving back and forth during capture, and thus defocus blur is provoked and strongly visible.
This can be achieved by manually turning the focus ring of the lens during recording.
The core algorithmic ingredient is a new video-based depth-from-defocus algorithm that computes space-time-coherent depth maps, deblurred all-in-focus video, and the focus distance for each frame.
We extensively evaluate our approach, and show that it enables compelling video post-processing effects, such as different types of refocusing.
\end{abstract}

\vspace{-1em}
\section{Introduction}
\label{sec:intro}
\vspace{-0.5em}

\begin{figure*}[t]
\includegraphics[width=\linewidth]{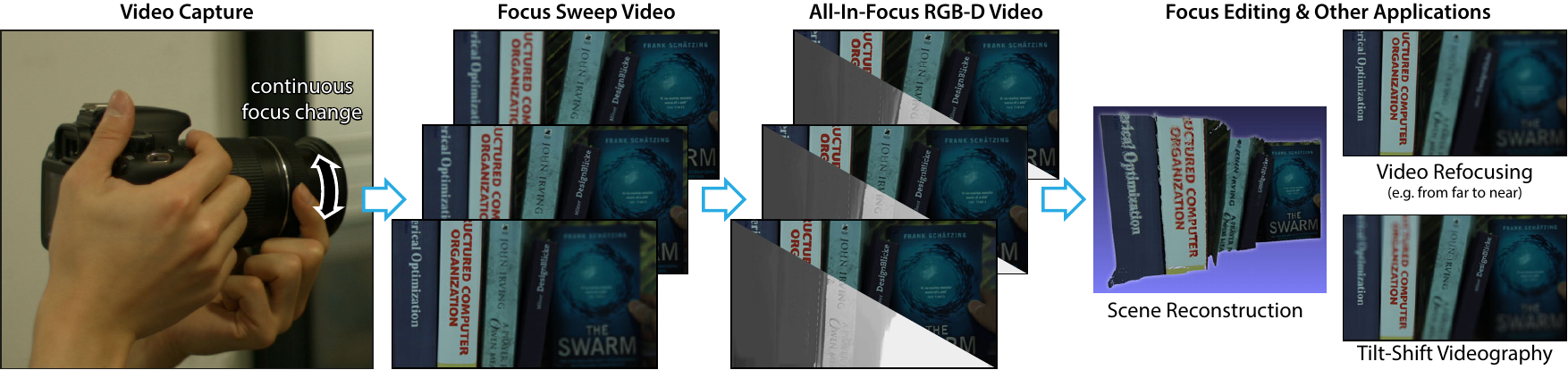} 
\caption{%
	\label{fig:teaser}%
	We capture focus sweep videos by continuously moving the focus plane across a scene, and then estimate per-frame \emph{all-in-focus RGB-D videos}.
	This enables a wide range of video editing applications, in particular video refocusing.
	Please refer to the paper's electronic version and supplemental video to more clearly see the defocus effects we show in our paper.
}\vspace{-1em}
\end{figure*}

Many cinematographically important video effects that normally require specific camera control during capture, such as focus and depth-of-field control, can be computationally achieved in post-processing if depth and focus distance are available for each video frame.
This post-capture control is on high demand by video professionals and amateurs alike.

In order to capture video and depth in general, and specifically to enable post-capture focus effects, several methods were proposed that use specialized camera hardware, such as active depth cameras \cite{RichaSDST2012}, light-field cameras \cite{NgLBDHH2005}, or cameras with coded optics \cite{LevinFDF2007}.

We follow a different path and propose one of the first end-to-end approaches for depth estimation from – and focus manipulation in – videos captured with an unmodified commodity consumer camera.
Our approach turns the often unwanted \textit{defocus blur}, which can be controlled by lens aperture, into a valuable signal.
In theory, smaller apertures produce sharp images for scenes covering a large depth range.
When using a larger aperture, only scene points close to a certain focus distance project to a single point on the image sensor and appear in focus (see \cref{sec:defocusmodel}).
Scene points at other distances are imaged as a \textit{circle of confusion} \cite{PotmeC1982}.
The limited region of sharp focus around the focus distance is known as \textit{depth of field}; outside of it the increasing \textit{defocus blur} is an important depth cue \cite{Mathe1996}.

Unfortunately, depth of field in normal videos cannot be changed after recording, unless a method like ours is applied.
Our approach (\cref{fig:teaser}) takes as input a video in which focus sweeps across the scene, e.g. by manual change of lens focus.
This means temporally changing defocus blur is purposefully provoked (see \cref{sec:videorecording}).
Each frame thus has a different focus setting, and no frame is entirely in focus.
Our core algorithmic contribution uses this provoked blur and performs space-time coherent \emph{depth}, \emph{all-in-focus color}, and \emph{focus distance} estimation at each frame.
We first segment the input video into multiple \textit{focus ramps}, where the focus plane sweeps across the scene in one direction.
The first stage of our approach (\cref{sec:alignment}) constructs a \textit{focus stack video} for each of them.
Focus stack videos consist of a focus stack at each video frame, by aligning adjacent video frames to the current frame using a new defocus-preserving warping technique.
At each frame, the focus stack video comprises multiple images with a range of approximately known focus distances (see \ifIncludeSuppMat{\cref{sec:init}}{Supplemental \cref*{supp-sec:init}}), which are used to estimate a depth map in the second stage (\cref{sec:depth}) using depth-from-defocus with filtering-based regularization.
The third stage (\cref{sec:deblurring}) performs spatially varying deconvolution to remove the defocus blur and produce all-in-focus images.
And the fourth stage of our approach (\cref{sec:focus}) further minimizes the remaining error by refining the focus distances for each frame, which significantly improves the depth maps and all-in-focus images in the next iteration of our algorithm.
Our end-to-end method requires no sophisticated calibration process for focus distances, which allows it to work robustly in practical scenarios.

In a nutshell, the main algorithmic contributions of our paper are:
	(1) a new hierarchical alignment scheme between video frames of different focus settings and dynamic scene content;
	(2) a new approach to estimate per-frame depth maps and deblurred all-in-focus color images in a space-time coherent way;
	(3) a new image-guided algorithm for focus distance initialization;
	(4) and a new optimization method for refining focus distances.
We extensively validate our method, compare against related work, and show high-quality refocusing, dolly-zoom and tilt-shift editing results on videos captured with different cameras.

\section{Related Work}
\label{sec:relatedwork}
\vspace{-0.5em}

\inlineheading{RGB-D Video Acquisition}
Many existing approaches for RGB-D video capture use special hardware, such as projectors or time-of-flight depth cameras~\cite{RichaSDST2012}, or use multiple cameras at different viewpoints.
%
Moreno-Noguer et al. \cite{MorenBN2007} use defocus blur and attenuation of the a projected dot pattern to estimate depth.
%
Coded aperture optics enable single-shot RGB-D image estimation~\cite{LevinFDF2007,BandoCN2008,LiangLWLC2008,ChakrZ2012,MartiWQLLWLKL2015}, but require more elaborate hardware modification.
%
Stereo correspondence \cite{BarroASH2015} or multi-view stereo approaches \cite{YuG2014} require multiple views, for instance by exploiting shaky camera motions.
%
Shroff et al. \cite{ShrofVTTAC2012} shift the camera's sensor along the optical axis to change the focus within a video.
They align consecutive video frames using optical flow to form a focus stack, and then apply depth from defocus to the stack.
Unlike all mentioned approaches, ours works with a single unmodified video camera without custom hardware.
There are also single-view methods based on 
non-rigid structure-from-motion \cite{RusseYA2014}, which interpret clear motion cues (in particular out-of-plane) under strong scene priors, and learning-based depth estimation methods \cite{HoiemEH2005,SaxenSN2009,SrivaSTTN2009,EigenF2015}.

\inlineheading{Depth from Focus/Defocus}
Focus stacking combines multiple differently focused images into a single \textit{all-in-focus} (or \textit{extended depth of field}) image \cite{PertuPGF2013}.
Depth-from-(de)focus techniques exploit (de)focus cues within a focus stack to compute depth.
Focus stacking is popular in macro photography, where the large lens magnification results in a very small depth of field.
By sweeping the focus plane across a scene or an object, each part of it will be sharpest in one photo, and these sharp regions are then combined into the all-in-focus image.
%
Depth from focus additionally determines the depth of a pixel from the focus setting that produced the sharpest focus \cite{Gross1987,NayarN1994}; however, this requires a densely sampled focus stack and a highly textured scene. 
Depth from defocus, on the other hand, exploits the varying degree of defocus blur of a scene point for computing depth from just a few defocused images \cite{Pentl1987,SubbaS1994}.
The all-in-focus image is then recovered by deconvolving the input images with the spatially-varying point spread function of the defocus blur.
Obviously, techniques relying on focus stacks only work well for scenes without camera or scene motion.

\citet{SuwajHS2015} proposed a hybrid approach that stitches an all-in-focus image using motion-compensated focus stacking, and then optimizes for the depth map using depth-from-defocus.
Their approach is completely automatic and even estimates camera parameters (up to an inherent affine ambiguity) from the input images.
However, their approach is limited to reconstructing a single frame from a focus stack, and cannot easily be extended to videos, as this requires stitching per-frame all-in-focus images.
Our approach is tailored for videos, not just single images.

\inlineheading{Refocusing Images and Videos}
Defocus blur is an important perceptual depth cue \cite{Mathe1996,HeldCOB2010}, thus refocusing is a powerful, appearance-altering effect.
However, just like RGB-D video capture, all approaches suitable for refocusing videos require some sort of custom hardware, such as special lenses \cite{NgLBDHH2005,MiauCN2013}, coded apertures \cite{LevinFDF2007} or active lighting \cite{MorenBN2007}.
Special image refocusing methods are difficult to extend to videos as they rely on multiple captures from the same view, for example, for depth from (de)focus \cite{SuwajHS2015}.

A single image captured with a light-field camera with special optics \cite{NgLBDHH2005,VeeraRAMT2007} can be virtually refocused \cite{IsaksMG2000,Ng2005}, but the light-field image often has a drastically reduced spatial resolution.
%
\citet{MiauCN2013} use a deformable lens to quickly and repeatedly sweep the focus plane across the scene while recording video at 120\,Hz.
They refocus video by selecting appropriately focused frames.
Some approaches exploit residual micro-blur that is hard to avoid even in a photograph set to be in focus everywhere. 
It can be used for coarse depth estimation \cite{ShiTXJ2015}, or removed entirely \cite{ShiXJ2015}.

Defocus deblurring is also related to motion deblurring \cite{ChoWL2012,HuXY2014,WulffB2014}. 
Their characteristics differ; motion blur is, for instance, mostly depth-independent.

\section{Preliminaries}
\label{sec:overview}
\vspace{-0.5em}

Both our lens model and aspects of video recording influence algorithm design.

\begin{figure}[h]
\centering\includegraphics[width=0.7\linewidth]{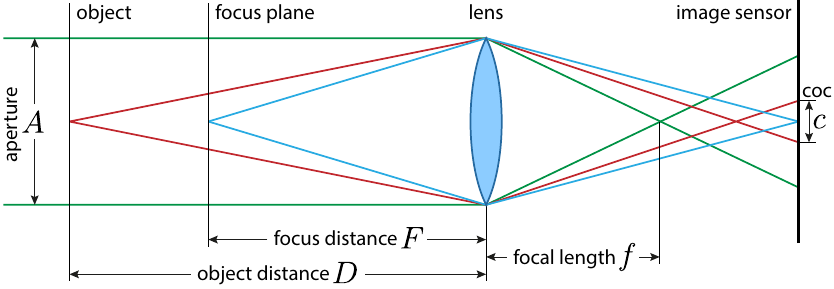}
\vspace{-0.25em}
\caption{%
	Thin-lens model and circle of confusion.
	}\vspace{-1em}
\label{fig:thinLens}
\end{figure}

\inlineheading{Defocus Model}
\label{sec:defocusmodel}
We assume a standard video camera with a finite aperture lens that produces a limited depth of field.
According to the thin-lens model (\cref{fig:thinLens}), the amount of defocus blur is quantified by the diameter $c$ of the circle of confusion \cite{PotmeC1982}:
\vspace{-0.5em}
\begin{align}
\label{eqn:coc}
c = \frac{A f \left| D - F \right|}{D(F - f)}  = \frac{f^2 | D - F |}{ND(F - f)} \text{,}
\end{align}
where $A \!=\! f/N$ is the diameter of the aperture, $f$ is the focal length of the lens, $N$ is the \textit{f}-number of the aperture, $D$ is the depth of a scene point and $F$ is the focus distance.
We assume fixed aperture and focal length, and a focus distance $F$ changing over time.
Therefore, the defocus blur of a 3D point only depends on its depth $D$ and the focus distance $F$, which we express as the point-spread function $K(D, F)$ corresponding to the circle of confusion in \cref{eqn:coc}.
We model the color of a defocused image $V$ at a pixel $\mathbf{x}$ using
\vspace{-0.25em}
\begin{align}
\label{eqn:defocus}
V(\mathbf{x}) = \left(K\!\left(D(\mathbf{x}), F\right) \ast I\right)\!(\mathbf{x}) \text{,}
\end{align}
where $I$ denotes the all-in-focus image.
Note that $K$ is spatially varying, because each pixel $\mathbf{x}$ may have a different depth $D(\mathbf{x})$.
For brevity, we hence omit the pixel index $\mathbf{x}$.

\inlineheading{Video Recording}
\label{sec:videorecording}
We record our focus sweep input video simply by manually adjusting the focus on the lens, roughly following a sinusoidal focus distance curve – leading to several focus ramps (see \cref{fig:overview}).
The exact focus distance for each video frame is unknown; reading it out from the camera is difficult in practice and may require low level modification of the firmware.
We use the \footurl{http://www.magiclantern.fm}{Magic Lantern} software for Canon EOS digital DSLR cameras to record timestamped lens information at about 4\,Hz.
In practice, focus distance values are not measured for most frames, timestamps may not be exactly aligned with the time of frame capture, and the recorded focus distances are quantized and not fully accurate.
There is also natural variation in the focus distance curves as people cannot exactly reproduce a curve.
Therefore, our algorithm uses the sparsely recorded lens information only as a guide and explicitly optimizes for the dense correct focus distances at every frame (\cref{sec:focus}).

\section{All-In-Focus RGB-D Video Recovery}
\label{sec:refocusing}
\vspace{-0.5em}

\begin{figure*}[t]
\centering
	\includegraphics[width=\linewidth]{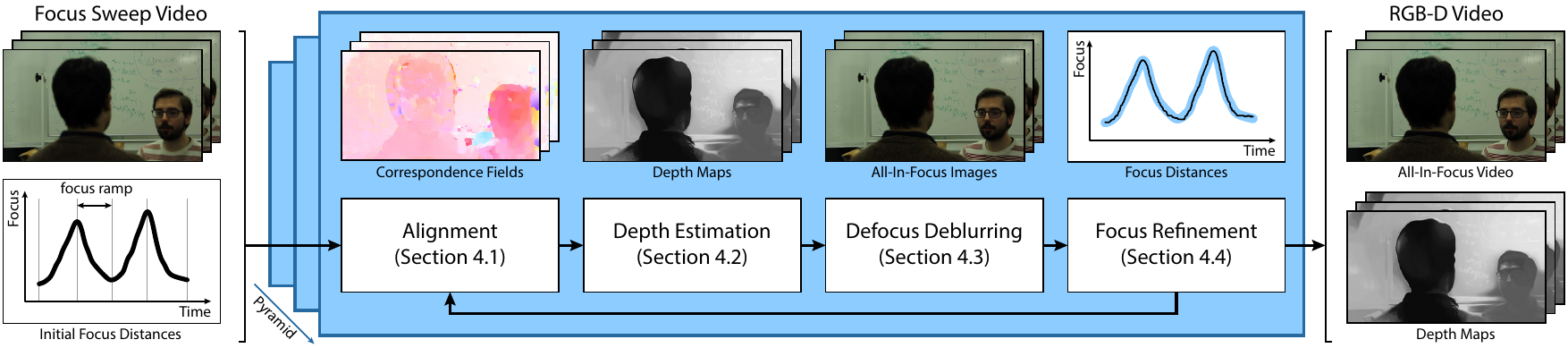} 
	\caption{%
		Overview of our approach.
		For each video frame, we first align neighboring frames to it to construct a focus stack.
		We then estimate spatially and temporally consistent depth maps from the focus stacks, and compute all-in-focus images using non-blind deconvolution.
		Finally, we refine the focus distances for all frames.
		We perform these steps in a coarse-to-fine manner and iterate until convergence.
	}\vspace{-1em}
	\label{fig:overview}
\end{figure*}

Given a video $\mathcal{V}$ with frames $\{V_t\}_{t \in T}$ containing one or more focus sweeps, we formulate our algorithm as a joint optimization framework that seeks the optimal depth maps $D_t$, all-in-focus images $I_t$, and focus distances $F_t$ for all frames $t \!\in\! T$.
Let us assume that $W_{s \rightarrow t}(\cdot)$ is a warping function that aligns an image at time~$s$ with time~$t$, while preserving the original defocus blur (explained in \cref{sec:alignment}).
Then, we can construct a focus stack at each frame $t$ by warping all input video frames to it using $\{ W_{s \rightarrow t}(V_s) \}_{s \in T}$ (in practice, we only warp keyframes, as explained later).
We seek the optimal depth map $D_t$ and all-in-focus image $I_t$, and focus distances $\{F_t\}_{t \in T}$ which best reproduce the focus stack at frame $t$ with the defocus model in \cref{eqn:defocus}.
$D_t$, $I_t$ and $F_t$, for $t \!\in\! T$, are the unknowns we solve for.
The core ingredient of our joint optimization is a data term that penalizes the defocus model error of the focus stack at all frames:
\vspace{-0.25em}\begingroup\makeatletter\def\f@size{8}\check@mathfonts
\begin{align}
\label{eqn:jointData}
E_\text{data} =
\sum_{t \in T}
\sum_{s \in T}
w_{t,s} \norm{ K(D_t, F_s) \ast I_t - W_{s \rightarrow t}(V_s) }^2 \text{.}
\end{align}%
\endgroup

\vspace{-0.25em}\noindent
We introduce the weighting term $w_{t,s}$ to give lower weights to pairs of frames that are further apart, and which hence need warping over longer temporal distances.
In our implementation, we use a Gaussian function $w_{t,s} \!=\! \exp(-\abs{t \!-\!s}^2 \!/ 2\sigma_\text{w}^2)$ with $\sigma_\text{w}$ set to 85 percent of the length of each focus ramp.

Simultaneously estimating depth, deblurring the input video and optimizing focus distances from purposefully defocused and temporally misaligned images is highly challenging; many invariance assumptions used by correspondence finding approaches break down in this case.
To solve this joint optimization problem efficiently, we decompose it into four subproblems, or \textit{stages}, that we solve iteratively:
defocus-preserving alignment (\cref{sec:alignment}),
depth estimation (\ref{sec:depth}),
defocus deblurring (\ref{sec:deblurring}),
and focus refinement (\ref{sec:focus}).
Initialization and implementation details can be found in \ifIncludeSuppMat{\cref{sec:init}}{Supplemental \cref*{supp-sec:init}}.
Each subproblem requires solving for a subset of the unknowns by minimizing 
\setlength{\columnsep}{8pt}
\begin{wrapfigure}[10]{r}{0.55\linewidth}
\vspace{-0.75\baselineskip}
\includegraphics[width=\linewidth]{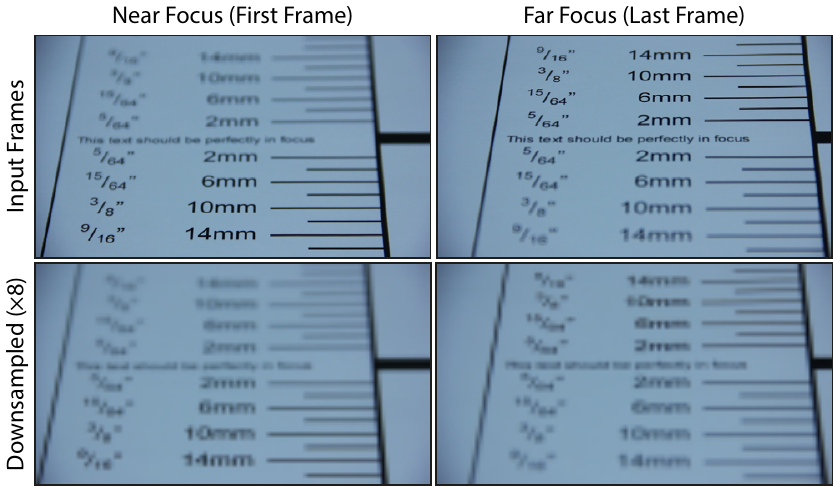}
\vspace{-1.25em}
\caption{%
	Downsampling (bottom) effectively reduces defocus difference, helping correspondence finding.
}
\label{fig:pyramid}
\end{wrapfigure}
a cost functional like \cref{eqn:jointData}, with additional regularization terms explained later.
We adopt a multi-scale, coarse-to-fine approach.
At each resolution level, we perform three iterations of the four stages, each of which is solved for the entire length of the input video.
The multi-resolution approach improves convergence, but more importantly, any focus difference between two video frames is reduced when the images are downsampled in the pyramid (see \cref{fig:pyramid}).
This insight enables us to compute reliable initial correspondence fields with less influence from different defocus blurs.
Once all parameters are estimated at a coarse level, the higher level of the pyramid uses them as initialization for its iterations.

\subsection{Patch-Based Defocus-Preserving Alignment}
\label{sec:alignment}

Here, we construct a focus stack for each frame of the input video using patch-based, defocus-preserving image alignment.
The result are the warping functions $W_{s \rightarrow t}$ for pairs ($s$, $t$) of frames, while all other unknowns ($I$, $D$ and $F$) remain constant.
Two frames in the focus sweep video, $V_s$ and $V_t$, generally differ in defocus blur and maybe scene or camera motion.
The main challenge of the defocus-preserving alignment is to compute a reliable correspondence field that is robust to both complex motion and defocus changes between the frames.

\begin{wrapfigure}[14]{r}{0.55\linewidth}
\vspace{-1.0\baselineskip}
\includegraphics[width=\linewidth]{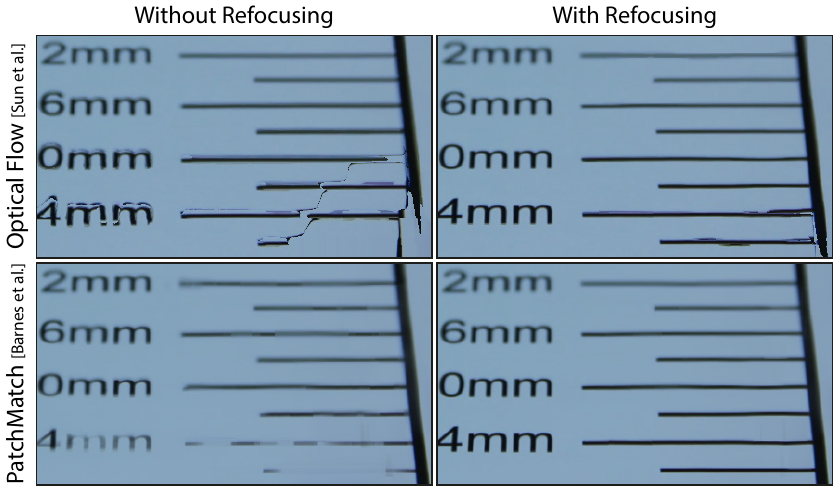}
\vspace{-1em}
\caption{%
	Comparison of focus stack alignment from near to far focus (\cref{fig:pyramid}).
	Without refocusing to match blur levels, both optical flow and PatchMatch fail.
	With refocusing, PatchMatch improves on flow (used by \citet{ShrofVTTAC2012}).
}
\label{fig:alignmentComparison}
\end{wrapfigure}
Using standard correspondence techniques, such as optical flow, to directly warp the input video frame $V_s$ to $V_t$ is prone to failure, because the different defocus blurs in the two images are not modeled by standard matching costs.
Optical flow will try to explain differences in defocus blur using flow displacements, which produces erroneous correspondences (see \cref{fig:alignmentComparison}).

The solution is to compensate any focus differences before computing correspondences \cite{ShrofVTTAC2012}.
We therefore refocus the target frame $V_t$ to match the focus distance $F_s$ of the source frame $s$ using the refocusing operator $ R(V_t, F_s) = K(D_t, F_s) \ast I_t $.
In the first iteration at the coarsest resolution level, the refocusing operator returns the input frame $V_t$ unchanged, as the downsampling in the pyramid has already removed most of the defocus blur.
In subsequent iterations, the refocusing operator uses the current estimates of frame $t$'s depth map $D_t$ and all-in-focus image $I_t$ to perform the refocusing.
The embedding of this focus difference compensation and alignment process into the overarching coarse-to-fine scheme enables reliable focus stack alignment even for large scene motions and notable defocus blur differences.

We use PatchMatch \cite{BarneSFG2009} to robustly compute the warping function $W_{s \rightarrow t}$ between the source frame $V_s$ and the refocused target frame $R(V_t, F_s)$.
PatchMatch handles complex motions and is fairly robust to the remaining focus differences, while traditional optical flow techniques tend to fail in such cases.
PatchMatch correspondences are not
\begin{wrapfigure}[8]{r}{0.55\linewidth}
\vspace{-1\baselineskip}
\includegraphics[width=\linewidth]{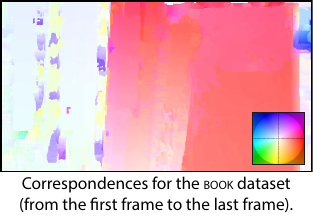}
\end{wrapfigure}
always geometrically correct (see right), as they exploit visually similar patches from other regions of the image which have similar defocus blur (note the purple and yellow regions on the left, which indicate vertical motion along the edge of the books).
In our case, this is an advantage, as it improves the warping quality while preserving defocus blur.
At the coarsest level of the pyramid, we initialize the PatchMatch search using optical flow \cite{SunRB2014}; at this level, focus-induced appearance differences are minimal.
We also constrain the size of the search window to find the best matches around the initial correspondences.
This encourages the estimated correspondence field to be more spatially consistent.
Since the warping is computed by refocusing the target frame, the defocus blur in the source frame is preserved, which is crucial for constructing valid focus stacks from a dynamic focus sweep video.

We apply the estimated defocus-preserving warping operators $W_{s \rightarrow t}$ to create a \textit{focus stack video} with per-frame focus stacks, as shown in \cref{fig:alignment}.
%
However, we do not warp all frames to all others, to prevent artifacts introduced by
\begin{wrapfigure}[16]{r}{0.55\linewidth}
\includegraphics[width=\linewidth]{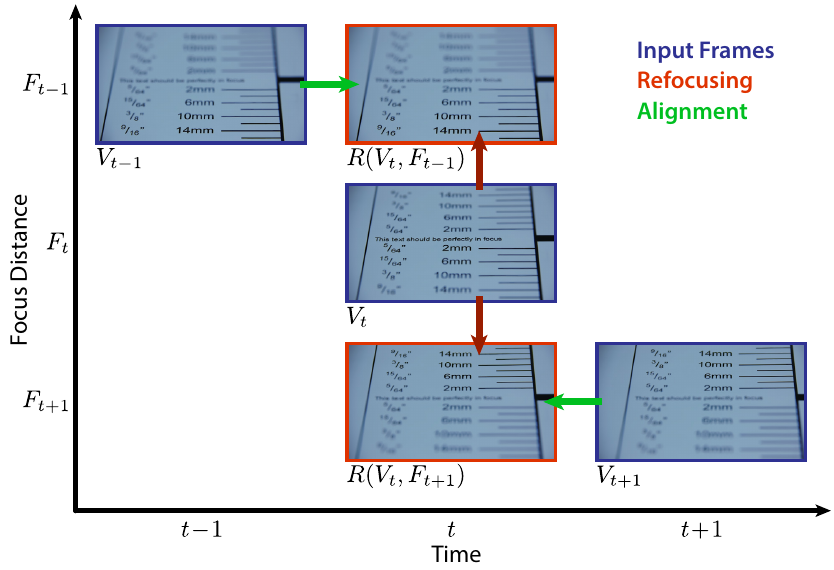}
\vspace{-1.0em}
\caption{%
		Defocus-preserving alignment.
		We refocus the input frame $V_t$ (center) to match the focus distances of neighboring frames $F_{t \pm 1}$ (red arrows), and compute correspondences (green arrows) between neighboring frames, $V_{t \pm 1}$, and the corresponding refocused image $R(V_t, F_{t \pm 1})$.
	}
\label{fig:alignment}
\end{wrapfigure}
aligning temporally distant videos frames in which the scene may have changed drastically.
Instead, we first segment the input video into contiguous focus ramps (see \cref{fig:overview}), $T_i \!\subset\! T$ for $i \!\in\! R$, which contain only temporally close frames.
For each input video frame $t$, we then create a focus stack by warping the other frames in its ramp to it using our defocus-preserving alignment.
This reduces the computational complexity of alignment from $\mathcal{O}(\abs{T}^2)$ for all-pairs warping, to $\mathcal{O}(\abs{T}^2 \!/\! \abs{R})$ for all-pairs warping within each focus ramp.

\subsection{Filtering-Based Depth Estimation}
\label{sec:depth}

The second stage of our approach estimates spatially and temporally consistent depth maps $D_t$ for all focus stacks, while keeping all other variables constant.
In this case, our data term $E_\text{data}$ from \cref{eqn:jointData} measures how well the estimated depth maps fit to the defocus observations in all focus stacks, which is equivalent to depth from defocus \cite{Pentl1987} applied per video frame.
This step requires the pixel-wise alignment across each focus stack, computed in the previous stage, to measure the fitting error.
Since this error is individually penalized at each pixel, it can lead to spatial inconsistencies in the depth map.
To avoid this issue, we introduce a long-range linear Potts model.
In contrast to the pairwise Potts model which compares depth values only between immediately adjacent pixels, our version performs long-range comparisons which benefit globally consistent depth estimation, yet prevents erroneous smoothing of actual features in the depth map:
\begingroup\makeatletter\def\f@size{8}\check@mathfonts
\begin{align}
\label{eqn:depthSmoothness}
	E_\text{smoothness}^\text{\ spatial} \!=
	\sum_{t \in T}
	\sum_\mathbf{x} \!
	\sum_{\mathbf{y} \neq \mathbf{x}}
	\min ( \alpha(\mathbf{x}, \mathbf{y}) \abs{D_t(\mathbf{x}) \!-\! D_t(\mathbf{y})}\!, \tau_d ) \text{,} 
\end{align}%
\endgroup%
where $\tau_d$ is the truncation value of the depth difference.
We use the bilateral weight $\alpha$ between two pixels $\mathbf{x}$ and $\mathbf{y}$,
$
\alpha(\mathbf{x}, \mathbf{y}) = \exp \! \left(
	-\frac{\norm{\mathbf{x} - \mathbf{y}}^2}{2\sigma_s^2}
	-\frac{\norm{I(\mathbf{x}) - I(\mathbf{y})}^2}{2\sigma_r^2}
\right) \!\text{,}
$
to encourage consistent depth estimation between nearby pixels with similar colors,
where $\sigma_s$ and $\sigma_r$ denote the standard deviation for the spatial and range terms, respectively.
We use $\sigma_s \!=\! \text{0.075} \!\times \text{the image width}$, and $\sigma_r \!=\! \text{0.05}$.

In addition, we want the depth maps to be temporally coherent across all frames.
We minimize the discrepancy between depth maps using
\begin{align}
\label{eqn:depthTemporal}
E_\text{smoothness}^\text{\ temporal} =
\sum_{t \in T}
\sum_{s \in T \backslash t} \norm{ D_t - W_{s \rightarrow t}(D_s) }^2 \text{,}
\end{align}
which encourages temporal consistency over extended depth map sequences.
The total cost function for depth map estimation is defined by combining the data term from \cref{eqn:jointData} and the smoothness terms from \cref{eqn:depthSmoothness,eqn:depthTemporal}:
\begin{align}
\label{eqn:depthTotal}
\argmin_{D} E_\text{data} + \lambda_\text{ss} E_\text{smoothness}^\text{\ spatial} + \lambda_\text{ts} E_\text{smoothness}^\text{\ temporal} \text{,}
\end{align}
where $\lambda_\text{ss} \!=\! \text{1}$ and $\lambda_\text{ts} \!=\! \text{0.2}$ are balancing  weights.

The direct minimization of \cref{eqn:depthTotal} requires global optimization with respect to all depth images, which is computationally expensive.
Instead, we solve an efficient approximation of the global optimization problem.
We pose the minimization task as a labeling problem, and first estimate spatially consistent depth maps for all frames by applying a variant of cost-volume filtering \cite{HosniRBRG2013}, and then refine the per-frame depth maps to enforce temporal consistency \cite{LangWASG2012}.

We start by computing per-frame depth maps $D_t$ in three steps.
(1) We evaluate the data term (\cref{eqn:jointData}) for $n$ pre-defined, uniformly spaced depth layers, and store the error for each pixel $\mathbf{x}$ and depth label $d$ in the cost volume $C(\mathbf{x}, d)$.
As in previous depth-from-defocus techniques \cite{Pentl1987}, we perform this evaluation in the frequency domain, where convolution can be efficiently computed using element-wise multiplication.
(2) We apply fast joint-bilateral filtering \cite{ParisD2009} on each depth-cost slice, to minimize the long-range spatial smoothness term in \cref{eqn:depthSmoothness}.
For this, we use the all-in-focus image $I_t$ as the guide image in computing the bilateral weight $\alpha$.
As in the previous section, we take the estimated all-in-focus image $I_t$ from the previous iteration, and assume $I_t \!=\! V_t$ in the first iteration of the coarsest resolution level.
(3) We select the spatially optimal depth for each pixel using
$D_t(\mathbf{x}) \!=\! \argmin_d \  C(\mathbf{x}, d)$.

\begin{wrapfigure}[10]{r}{0.59\linewidth}
\vspace{-0.75\baselineskip}
\includegraphics[width=\linewidth]{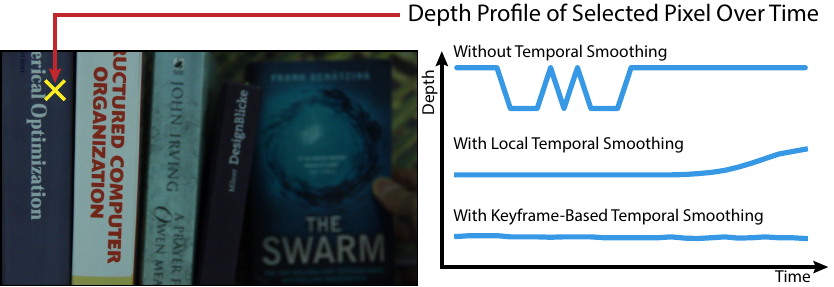}
\vspace{-1em}
\caption{%
	Our keyframe-based smoothing produces more consistent depth than local smoothing of adjacent frames.
	(The pixel should have constant depth in this scene.)
}
\label{fig:temporalConsistencyEvaluation}
\end{wrapfigure}
After computing depth maps independently from each focus stack, we apply temporal smoothing to make the depth maps consistent over time.
We use a keyframe-based approach with a sliding temporal window.
For each frame $t$, we align the depth maps of the previous and following two keyframes to the current depth map $D_t$ using our warping operator $W_{s \rightarrow t}$ computed on the all-in-focus images.
The updated depth map $D_t$ is the Gaussian-weighted mean of aligned depth maps.
The used keyframes are not restricted to be from the same focus ramp as the frame $t$; this enforces temporal consistency also across focus ramp boundaries.
In \cref{fig:temporalConsistencyEvaluation}, we show that our approach successfully produces temporally coherent depth maps, compared to the unfiltered input depth maps and also the simpler local filtering of adjacent frames, as some bias remains due to the short temporal range of the filtering.

\subsection{Defocus Deblurring}
\label{sec:deblurring}

\setlength{\columnsep}{8pt}
\begin{wrapfigure}[9]{r}{0.62\linewidth}
\vspace{-1.1\baselineskip}
\includegraphics[width=\linewidth]{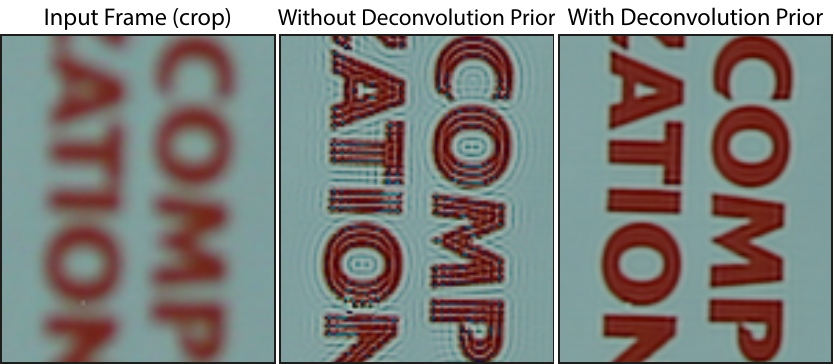}
\vspace{-1em}
\caption{%
		Deconvolution can result in ringing artifacts, which are suppressed by our deconvolution smoothness prior.
	}
\label{fig:deconvolutionComparison}
\end{wrapfigure}
Now that we have computed the depth maps $D_t$, we estimate all-in-focus images $I_t$ using non-blind deconvolution with the spatially varying point-spread function (PSF) corresponding to the depth-dependent circle of confusion (\cref{eqn:coc}).
While the disc shape of the PSF is a good approximation of the actual shape of the camera aperture, in practice, its sharp boundary causes ringing artifacts in the deconvolution process due to zero-crossings in the frequency domain \cite{LevinFDF2007}, see \cref{fig:deconvolutionComparison}.
We therefore adopt the smoothness term introduced by \citet{ZhouLN2011} to prevent ringing artifacts:
\begin{align}
\label{eqn:deblurringSmoothness}
E_\text{smoothness}^\text{\ all-in-focus} = \sum_{t \in T} \| H \ast I_t \|^2,
\end{align}
where $H$ is an image statistics prior.

The smoothness term uses a learning approach to capture natural image statistics.
We first take sample images at the same resolution as the input image from a database of natural images, and then apply the Fourier transform to the samples to compute the frequency distribution of image statistics.
The final image statistics of the natural image dataset $H$ is obtained by averaging the squared per-frequency modulus of all sample distributions.
The smoothness term \cref{eqn:deblurringSmoothness} enforces the all-in-focus image $I_t$ to follow a similar frequency distribution as the learned one in $H$.
The image statistics prior $H$ only needs to be computed once for each video resolution to be processed, and can then be reused for new videos.
For the details of the smoothness term, we refer the reader to \citet{ZhouLN2011}.

The total cost function of the defocus deblurring is a combination of the data term in \cref{eqn:jointData} and the learned smoothness term in \cref{eqn:deblurringSmoothness}:
\begin{align}
\label{eqn:deblurringTotal}
\argmin_{I} E_\text{data} + \lambda_\text{as} E_\text{smoothness}^\text{\ all-in-focus} \text{,}
\end{align}
where $\lambda_\text{as} \!=\! 10^{-3}$ balances the two cost terms.
We compute the optimal all-in-focus image $I_t$ by performing Wiener deconvolution independently on a range of $n$ depth layers, each with a fixed, depth-dependent point spread function, and then composite the sub-images to obtain the all-in-focus image $I_t$.

\subsection{Focus Distance Refinement}
\label{sec:focus}
\vspace{-0.5em}

This step refines the focus distances to reduce the data term $E_\text{data}$ (\cref{eqn:jointData}).
As explained in \cref{sec:videorecording}, we can at best read out temporally sparse focus distance values from the camera, which are moreover subject to inaccuracies.
To overcome this difficulty, we refine the focus distances for all frames in the final stage of our approach.
By rearranging the terms associated with the focus distance $F_t$ in \cref{eqn:jointData}, we define the focus refinement subproblem as
\vspace{-0.5em}
\begin{align}
\label{eqn:focusDistanceRefinement}
\argmin_{F_t} \sum_{s \in T} w_{s,t} \norm{ R(V_s, F_t) - W_{t \rightarrow s}(V_t) }^2 \text{.}
\end{align}
\vspace{-0.75em}
\setlength{\columnsep}{8pt}
\begin{wrapfigure}[15]{r}{0.55\linewidth}
	\vspace{-1.25\baselineskip}
	\includegraphics[width=\linewidth]{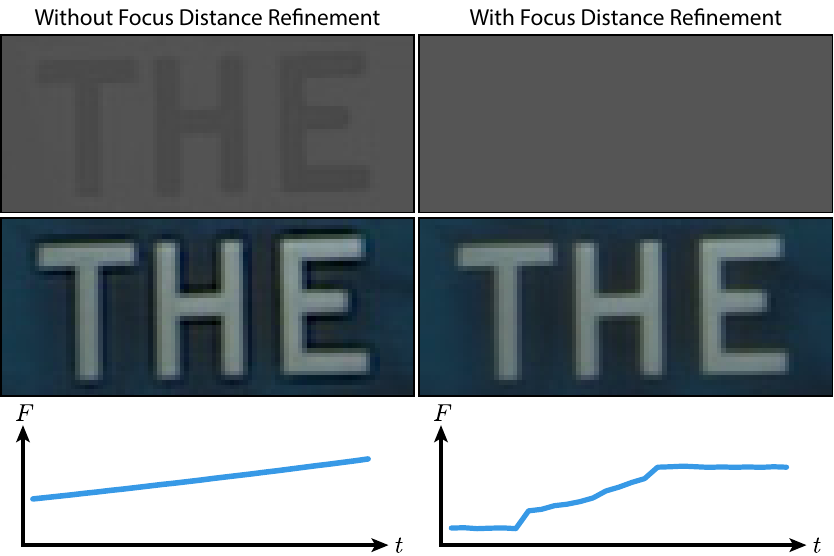}
	\caption{Focus distance refinement improves depth maps (top) by reducing texture copy artifacts, and also removes halos in all-in-focus images (middle).
	The refined focus distances (bottom) correctly reflect the constant focus at the beginning and end of the video.
	}
	\label{fig:focusOptimizationExample}
\end{wrapfigure}
We optimize this equation by gradient descent.
Since the cost function is highly nonlinear in $F_t$, we compute the gradient numerically by examining the costs for focus distances $F_t \!\pm\! \delta$ with $\delta \!=\! \text{5\,mm}$.
In practice, we refocus each source frame $V_s$ to focus distances $F_t \!\pm\! \delta$, and compare it to the aligned target frame $W_{t \rightarrow s}(V_t)$ (see \ifIncludeSuppMat{\cref{fig:focusOptimization}}{Supplementary \cref*{supp-fig:focusOptimization}}).
We then set the focus distance $F_t$ to the new minimum.

We demonstrate the performance of our focus distance refinement in \cref{fig:focusOptimizationExample}.
It improves the depth estimation as well as visual quality of the all-in-focus images by suppressing excessive edge contrasts.
Because this strategy frees us from requiring artificial patterns or special hardware for the accurate calibration of focus distances, it allows for our flexible and simple acquisition of the focus sweep video.

\vspace{-0.25em}
\section{Results and Evaluation}
\label{sec:results}
\vspace{-0.5em}

\begin{figure}[htp!]
\centering
\includegraphics[width=0.95\linewidth]{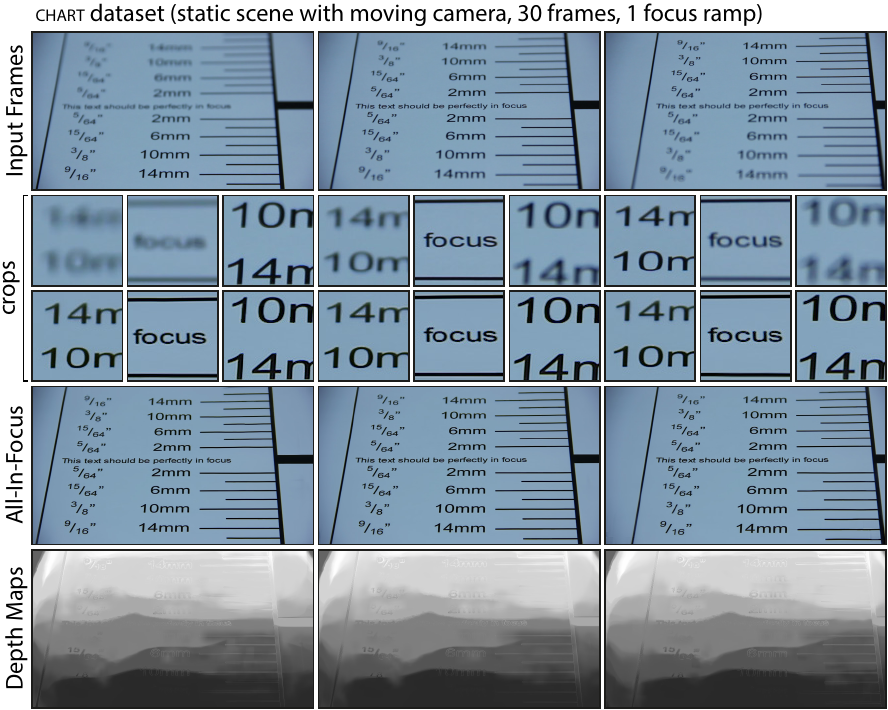}\vspace{0.5em}
\includegraphics[width=0.95\linewidth]{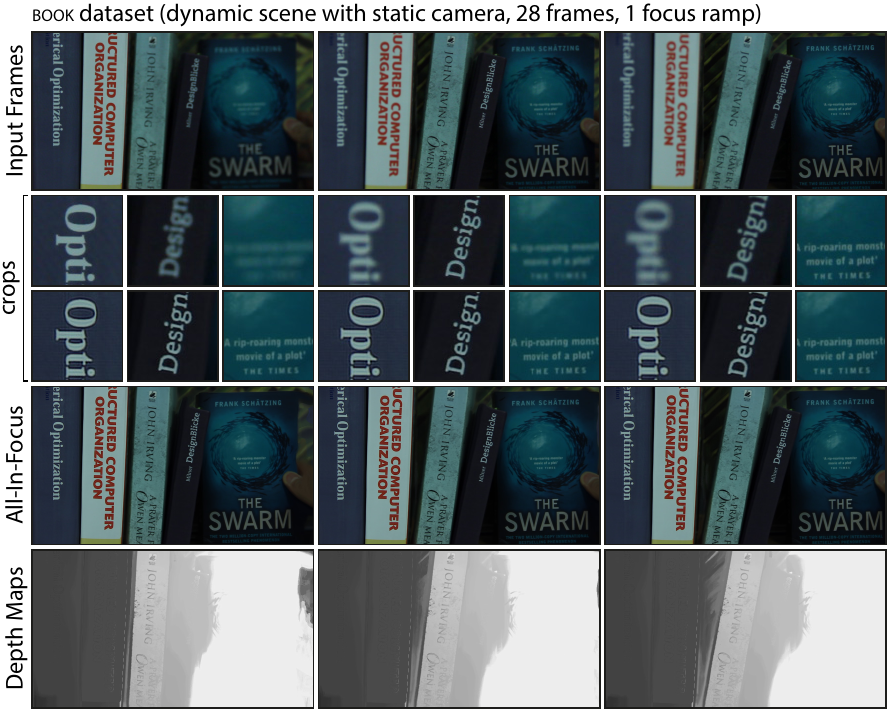}\vspace{0.5em}
\includegraphics[width=0.95\linewidth]{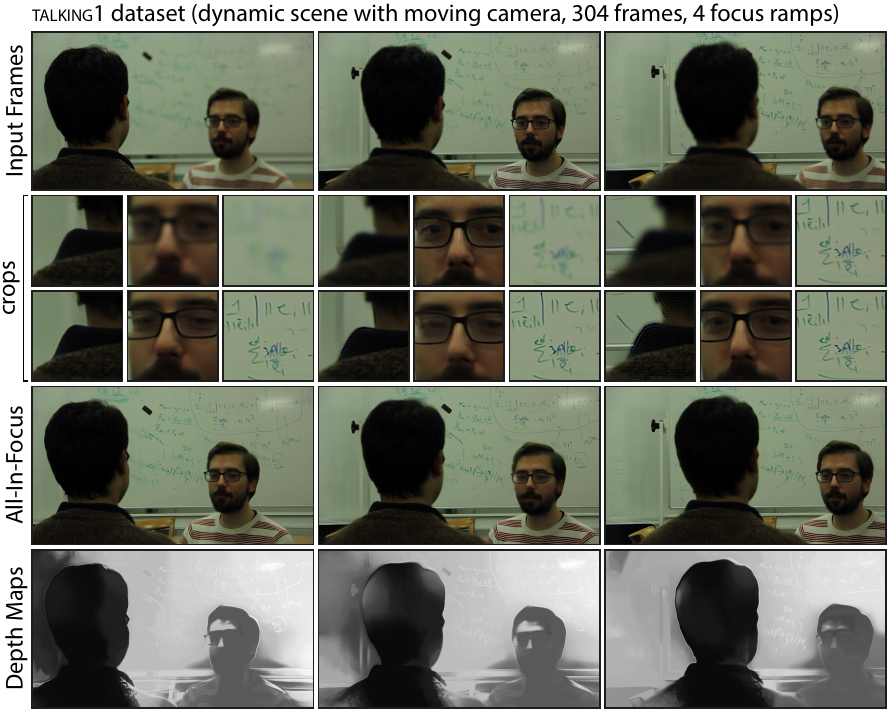}
\caption{
	RGB-D video results.
	We show reconstructed all-in-focus images and depth maps for three focus sweep videos with various combinations of scene and camera motion.
	The image crops (top: input frame cropped, bottom: all-in-focus images cropped) focus on regions at the near, middle and far end (from left to right) of the scene's depth range.
	Note that each input frame is in focus in only one of the three crops, while our all-in-focus images are in focus everywhere. Please zoom in to see more details.
}
\label{fig:results}
\end{figure}

\begin{figure}[htp!]
	\centering
	\includegraphics[width=\linewidth]{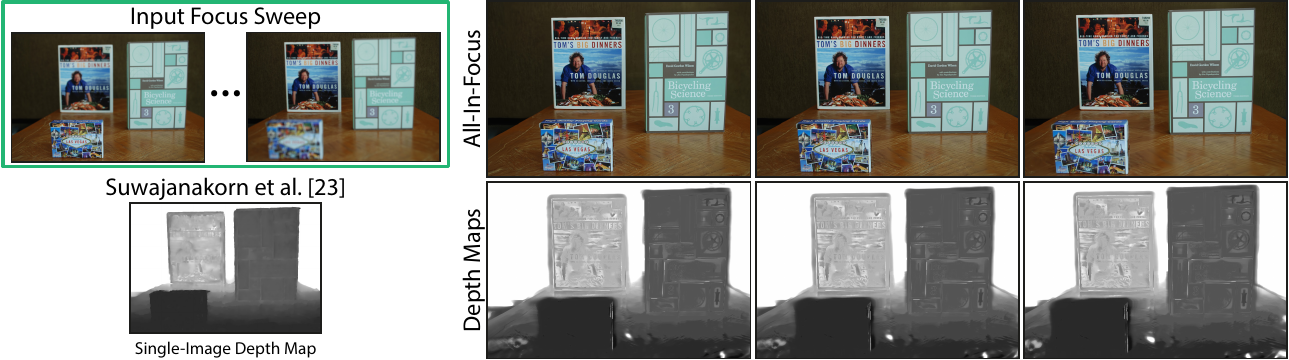}
	\caption{
		Comparison of our approach to Suwajanakorn et al.~\cite{SuwajHS2015} on their dynamic dataset.
		Left: Input focus stack, focused near (left) to far (right), and their estimated depth map for the last frame only.
		Right: We reconstruct all-in-focus images and depth maps for all frames of this dynamic sequence (also see our video).
	}\vspace{-1.5em}
	\label{fig:suwajanakorn}
\end{figure}

We thoroughly evaluate our proposed video depth-from-defocus approach for reconstructing all-in-focus RGB-D videos.
We first show qualitative results on natural, dynamic scenes with non-trivial motion, captured with static and moving video cameras.
We then compare our approach against the two closest approaches, by \citet{ShrofVTTAC2012} and \citet{SuwajHS2015}.
We further evaluate the design choices made in our approach with an ablation study on a ground-truth dataset.
Lastly, we evaluate our focus refinement optimization in \ifIncludeSuppMat{\cref{sec:focus_refinement_evaluation}}{Supplemental \cref*{supp-sec:focus_refinement_evaluation}}.

We show all-in-focus images and depth map results on a range of datasets in \ifIncludeSuppMat{\cref{fig:results,fig:results-supp},}{\cref{fig:results}, in Supplemental \cref*{supp-fig:results-supp}} and in our video.
Our depth maps capture the gist of each scene, including the main depth layers and their silhouettes, and the depth gradients of slanted planes with sufficient texture.
As demonstrated by the results, our approach works for dynamic scenes, and handles a fair degree of occlusions, dis-occlusions and out-of-plane motions.
It also properly reconstructs the depth and all-in-focus appearance of small objects, like the earrings in sequence \textsc{talking2} (\ifIncludeSuppMat{\cref{fig:results-supp}}{Supplemental \cref*{supp-fig:results-supp}}), which is highly challenging.
Note that our approach also works if scene and camera are rather static, where approaches requiring notable disparity for depth estimation would fail – even on unblurred footage. 
Similar to previous depth-from-(de)focus techniques, our approach works best for textured scenes that are captured in a full focus stack.
Although our depth maps are not perfect, they are temporally coherent and enable visually plausible video refocusing applications\ifIncludeSuppMat{ (see \cref{sec:applications})}{, as shown in Supplemental \cref*{supp-sec:applications}}.

\inlineheading{Comparison to \citet{ShrofVTTAC2012}}
This work moves a camera's sensor along the optical axis to compute all-in-focus RGB-D videos in an approach similar to ours.
However, our approach improves on theirs in several important ways:
(1) we use a commodity consumer video camera that does not require any hardware modifications like in their approach,
(2) our defocus-preserving alignment finds more reliable correspondences than optical flow,
(3) our depth maps are more detailed and temporally coherent, and
(4) our all-in-focus images and hence refocusing results improve on theirs.
We simulate their focus stack alignment approach by replacing PatchMatch in our implementation with optical flow \cite{SunRB2014}.
\Cref{fig:alignmentComparison} shows that PatchMatch achieves visually better alignment results.

\begin{figure*}[htp!]
\centering
\includegraphics[]{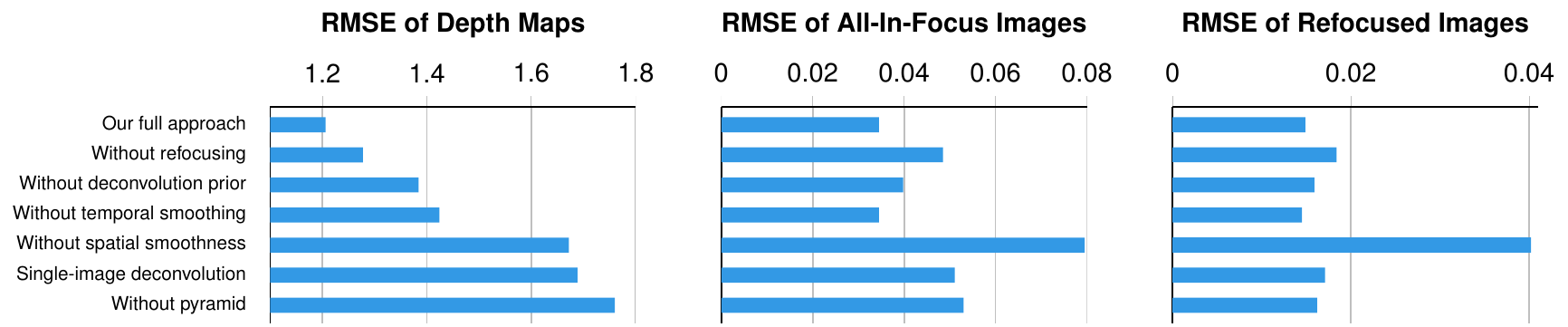}
\caption{
	Validation of design choices using an ablation study
	(lower RMSE is better).
	Our approach is best overall, but each components is required for achieving best results.
\vspace{-1.25em}
}
\label{fig:ablation}
\end{figure*}

\inlineheading{Comparison to \citet{SuwajHS2015}}
This recent depth-from-focus technique computes a \emph{single} depth map with all-in-focus image from quick focus sweeps of around 30 photos of static scenes with little camera motion.
Their approach first reconstructs the all-in-focus image by aligning the input photos and stitching the sharpest regions.
This will fail for videos, as dynamic scenes break their alignment strategy of concatenating the optical flows. 
Any estimated per-frame depth maps are also most likely not temporally coherent.
Our approach, on the other hand, computes temporally coherent all-in-focus RGB-D videos of dynamic scenes.
Our robust defocus-preserving alignment (\cref{sec:alignment}) enables us to construct per-frame focus stacks for dynamic scenes (moving scene and camera), and hence to compute per-frame depth maps and all-in-focus images.
On top, we implement keyframe-based temporal consistency filtering (\cref{sec:depth}) to remove any flickering from the depth maps.
We visually compare our results to theirs on one of their datasets in \cref{fig:suwajanakorn}.
We use their provided camera parameters without further focus distance refinement (\cref{sec:focus}).

\inlineheading{Validation of Design Choices}
We performed a quantitative ablation study to analyze the influence of the design choices in our algorithm.
For this, we synthetically defocus 10 frames from the MPI-Sintel dataset `alley\_1' \cite{ButleWSB2012} using two focus ramps, and apply additive Gaussian noise with $\sigma \!=\! \text{3}/\text{255}$ to simulate camera imaging noise.
We then process the resulting video while disabling or replacing individual components of our approach.
In \cref{fig:ablation}, we evaluate the accuracy of the estimated depth maps and all-in-focus images using the root-mean-squared error (RMSE) compared to the ground truth.
Our full approach produces overall the best results.
One can clearly see the importance of each component in our approach, as leaving them out significantly degrades the quality of the estimated depth maps or all-in-focus images, or both.
We also evaluate how accurately each alternative explains the input defocus images when refocusing the all-in-focus image using the estimated depth map.
Without temporal smoothing, refocused images have lower RMSE than our full approach, but the images lack temporal consistency (which is not measured by RMSE).

\section{Discussion and Conclusion}
\label{sec:discussion}
\vspace{-0.25em}

\inlineheading{Limitations}
Our approach relies on aligning all frames of a focus ramp to each other.
This works well for focus ramps of up to around 30 frames, but becomes more difficult for longer ramps, as more motion needs to be compensated.
This is significantly more difficult than for example the alignment required for HDR video reconstruction~\cite{KalanSBDGS2013}, which only needs to align three subsequent frames instead of 30–100.
While our alignment approach produces good results within one ramp, even a long one, the consistency across long ramps becomes more difficult to enforce.
This may lead to popping artifacts in the all-in-focus video.

As in previous depth-from-defocus methods, each aligned focus stack yields a single depth map.
However, this limits objects in adjacent video frames to have similar depths, which restricts their motion in depth.

\noindent
Large occlusions are also problematic as the focus stack alignment degrades in quality when part of the scene is not visible during a focus ramp, for example at the image boundaries.
Like static depth-from-defocus methods, we assume the appearance of objects as well as lighting remain constant.
Additionally, untextured regions are harder to reconstruct, and may show some temporal flickering, similar to previous depth-from-defocus methods but also passive, image-based depth reconstruction approaches in general.

We employ a simple blur model, which uses a spatially varying convolution with a point-spread function.
This may cause blurs across depth boundaries, which can create halos in the depth maps (see discussion in Lee et al. \cite{LeeES2010}).
A potential solution are more sophisticated, multi-layer defocus blur models \cite{KrausS2007}, which are harder to integrate into our optimization.
Our depth maps are plausible.
They may not entirely match the quality of depth maps from RGB-D cameras or multi-camera systems, but they were recorded with a completely unaltered camera, along with focus distances and all-in-focus frames, and enable video focus post-processing at good quality.
Our goal was to explore what is possible with unaltered hardware and what information may lie in typical artifacts, even auto-focus pulls.   


\inlineheading{Conclusion}
\label{sec:conclusion}
We presented the first algorithm for space-time coherent depth-from-defocus from video. 
It reconstructs all-in-focus RGB-D video of dynamic scenes with an unmodified commodity video camera.
We open a different view on RGB-D video capture by turning the often unwanted defocus blur artifact into a valuable signal. 
From an input video with purposefully provoked defocus blur, e.g. by simply turning the lens, we compute space-time-coherent depth maps, deblurred all-in-focus video and per-frame focus distance.
Our end-to-end approach relies on several algorithmic contributions, including an alignment scheme robust to strongly varying focus settings, an image-based method for accurate focus distance estimation, and a space-time-coherent depth estimation and deblurring approach. 
We have extensively evaluated our method and its components, and show that it enables compelling video refocusing effects.

\inlineheading{Acknowledgements}
We thank the authors of the used datasets.
Funded by ERC Starting Grant 335545 CapReal.

\ifIncludeSuppMat{
\appendix
\section{Initialization and Implementation}
\label{sec:init}

The input to our video depth-from-defocus approach is a radiometrically linearized video with temporally changing focus distances containing one or more focus ramps, but with otherwise constant camera settings.
We assume that we know camera properties such as the focal length, aperture \textit{f}-number, sensor size, as well as temporally sparse readings of the camera's focus distances for some video frames.

\inlineheading{Focus Distance Initialization}
Before the start of our algorithm in \ifIncludeSuppMat{\cref{sec:refocusing}}{\cref*{main-sec:refocusing}} of the main paper, we compute an initial set of temporally dense focus distance values.
We use the sparse timestamped focus distance readings from the Magic Lantern firmware as starting point, see \ifIncludeSuppMat{\cref{sec:videorecording}}{\cref*{main-sec:videorecording}} in the main paper.
We then solve for the per-frame focus distances $F$ using an energy minimization with the recorded focus data as data term, and additional smoothness and focus-consistency regularization terms:
\begin{equation}
\label{eqn:focus_init_energy}
\argmin_F E_\text{data}^\text{focus} + \lambda_\text{fs} E_\text{smoothness}^\text{focus} + \lambda_\text{focus} E_\text{focus} \text{.}
\end{equation}
The recorded focus distances $F_t^\text{rec}$ are available only for some frames $t \!\in\! T_\text{rec}$, so we constrain the unknown focus distances $F_t$ at those frames to lie close to them:
\begin{equation}
\label{eqn:focus_init_data}
E_\text{data}^\text{focus} = \sum_{t \in T_\text{rec}} \norm{ F_t - F_t^\text{rec} }^2 \text{.}
\end{equation}
As the focus is assumed to change smoothly over time, we enforce this by penalizing the second derivative of the focus distances:
\begin{equation}
\label{eqn:focus_init_smooth}
E_\text{smoothness}^\text{focus} = \sum_t \norm{ F_{t - 1} - 2 F_t + F_{t + 1} }^2 \text{.}
\end{equation}
The focus-consistency term exploits the observation that similar focus distances result in similar depth-of-field and hence similar images, so if video frames appear very similar, then their focus distances should also be similar (see \cref{fig:FocusRefinement}):
\begin{equation}
\label{eqn:focus_init_focus}
E_\text{focus} = \sum_t \sum_{s \neq t} s_{t,s} \norm{ F_t - F_s }^2 \text{,}
\end{equation}
where $s_{t,s}$ measures the (symmetric) similarity of the input video frames $V_t$ and $V_s$, so that more similar frames enforce consistency constraints more strongly.
We compute the similarity using
\begin{equation}
\label{eqn:focus_init_sim}
s_{t,s} = \min\!\left(0, 1 - \min\!\left( d_{t,s}, d_{s,t} \right) / \tau_\text{sim} \right) \text{,}
\end{equation}
based on the image dissimilarity $d_{t,s}$ which we compute using the RMSE between input frames $V_t$ and $V_s$ warped to $V_t$ using low-resolution (160$\times$90) optical flow to compensate for camera and scene motion.
The similarity threshold $\tau_\text{sim}$ determines which pairs of input frames result in consistency constraints and how strongly they are enforced.
Typical parameter values are $\lambda_\text{fs} \!=\! \sqrt{10}$, $\lambda_\text{focus} \!=\! 0.1$ and $\tau_\text{sim} \!\in\! [0.01, 0.05]$.

\begin{figure}[t]
	\centering
	\includegraphics[width=\linewidth]{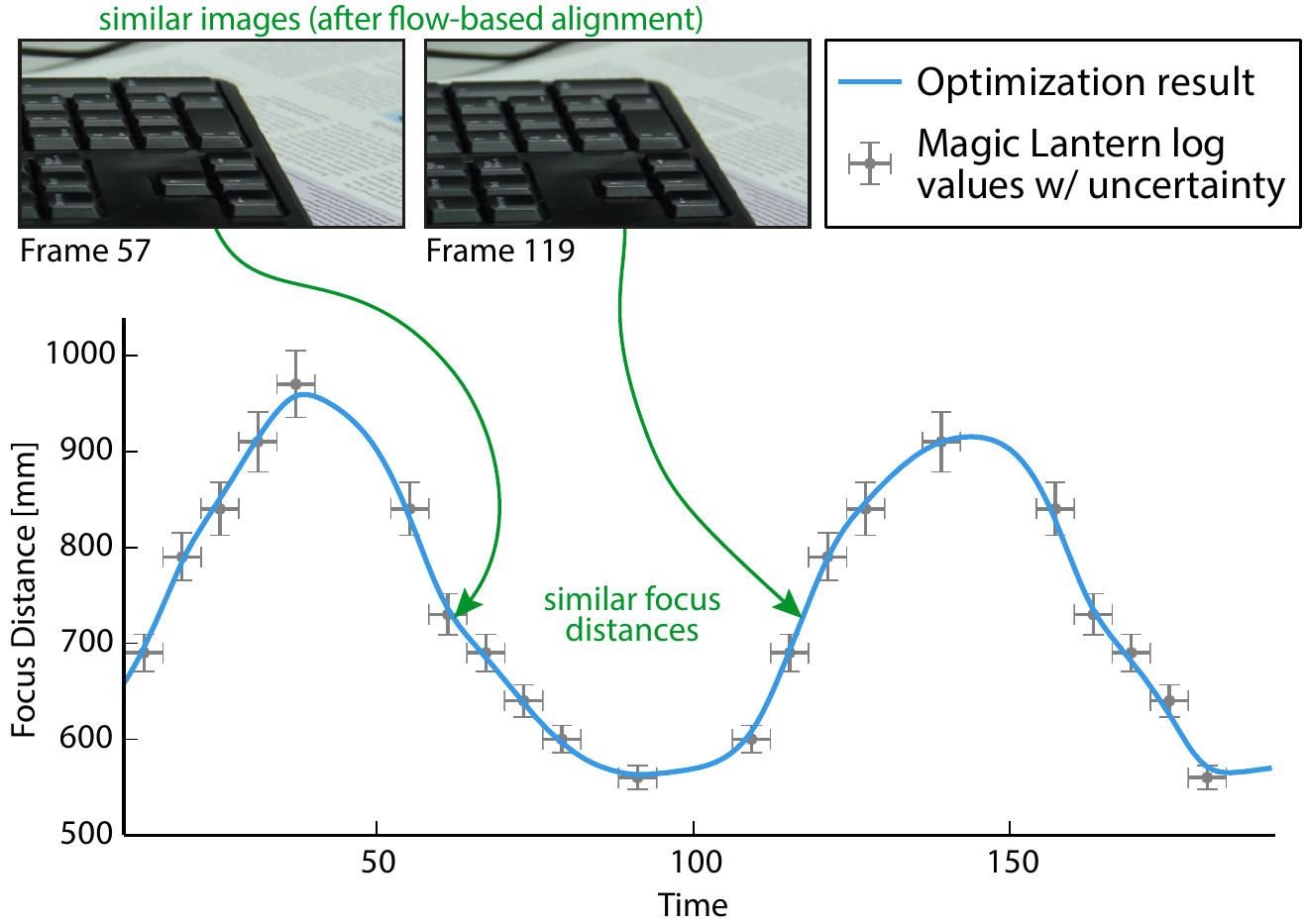}
	\caption{
		Focus distance initialization yields a smooth initial focus curve from sparse Magic Lantern data.
		Similar images according to $s_{t,s}$ (\cref{eqn:focus_init_sim}) enforce consistent focus distances.
	}\vspace{-1em}
	\label{fig:FocusRefinement}
\end{figure}

\subsection{Implementation of Video Depth-From-Defocus Algorithm}

To implement our method from \ifIncludeSuppMat{\cref{sec:refocusing}}{\cref*{main-sec:refocusing} (main paper)}, we use a multi-resolution approach with three levels to improve the convergence and visual quality of our results, as image defocus is more similar at coarser image resolutions.
At each pyramid level, we perform three iterations of the stages described in \ifIncludeSuppMat{\cref{sec:alignment,sec:depth,sec:deblurring,sec:focus}}{\cref*{main-sec:alignment,main-sec:depth,main-sec:deblurring,main-sec:focus} of the main paper}.
At the coarsest level, we start the first iteration assuming that the all-in-focus image $I_t$ is the input video frame $V_t$, 
and also initialize our PatchMatch correspondences using optical flow \cite{SunRB2014} to provide a good starting point for our alignment computations.
Between pyramid levels, we bilinearly upsample the all-in-focus images $I_t$, depth maps $D_t$ and all computed flow fields.
We use scale-adjusted patch sizes for PatchMatch, using 25$\times$25 pixels at the finest level and 7$\times$7 at the coarsest.

\inlineheading{Computation Times}
Our all-in-focus RGB-D video estimation approach processes 30 video frames at 854$\times$480 resolution in 4 hours on a 30-core 2.8\,GHz processor with 256 GB memory. 
This runtime breaks down as follows, per video frame:
8.6 minutes for defocus-preserving alignment,
19 minutes for depth estimation,
2.4 minutes for defocus deblurring, and
5 seconds for focus distance refinement.
Our \textsc{matlab} implementation is unoptimized, but parallelized over the input video frames.
We believe an optimized, possibly GPU-assisted implementation would yield significant speed-ups.

\section{Results}
\label{sec:results-supp}
We show additional all-in-focus images and depth map results on a range of datasets in \cref{fig:results-supp}.

\begin{figure}[htp!]
\centering
\includegraphics[width=0.95\linewidth]{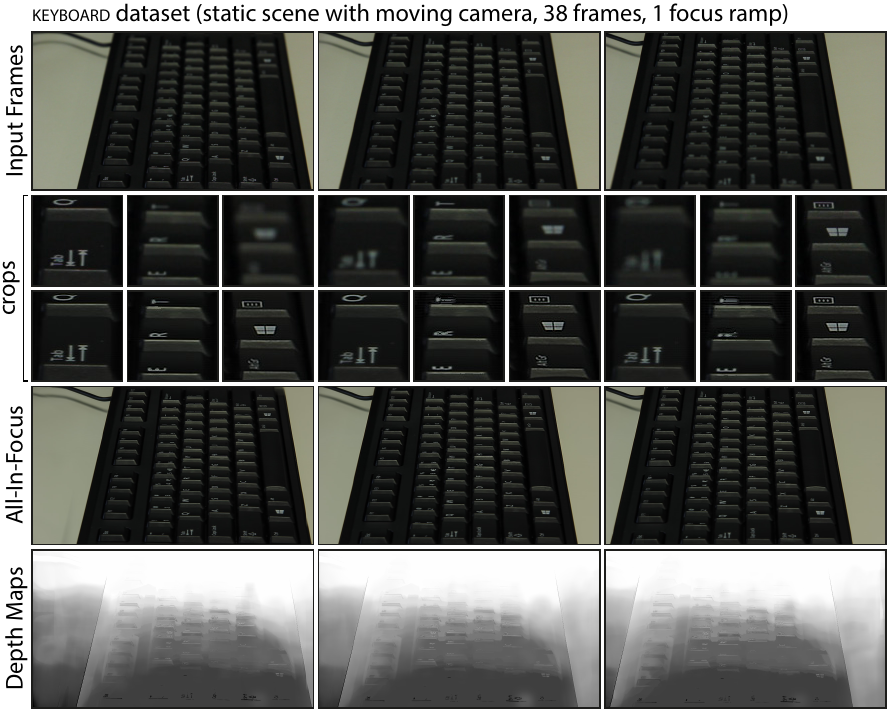}\vspace{0.5em}
\includegraphics[width=0.95\linewidth]{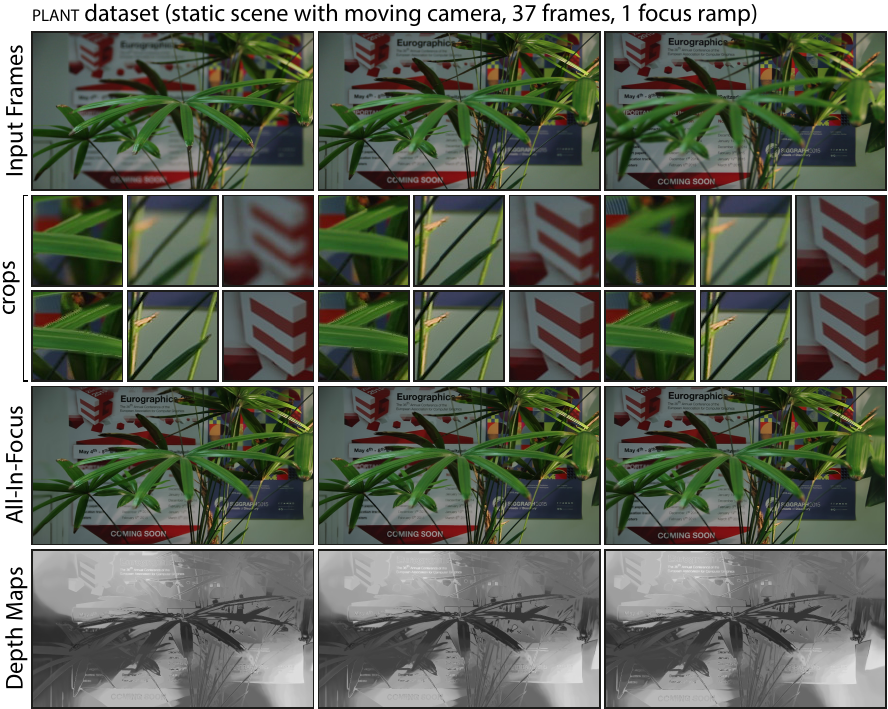}\vspace{0.5em}
\includegraphics[width=0.95\linewidth]{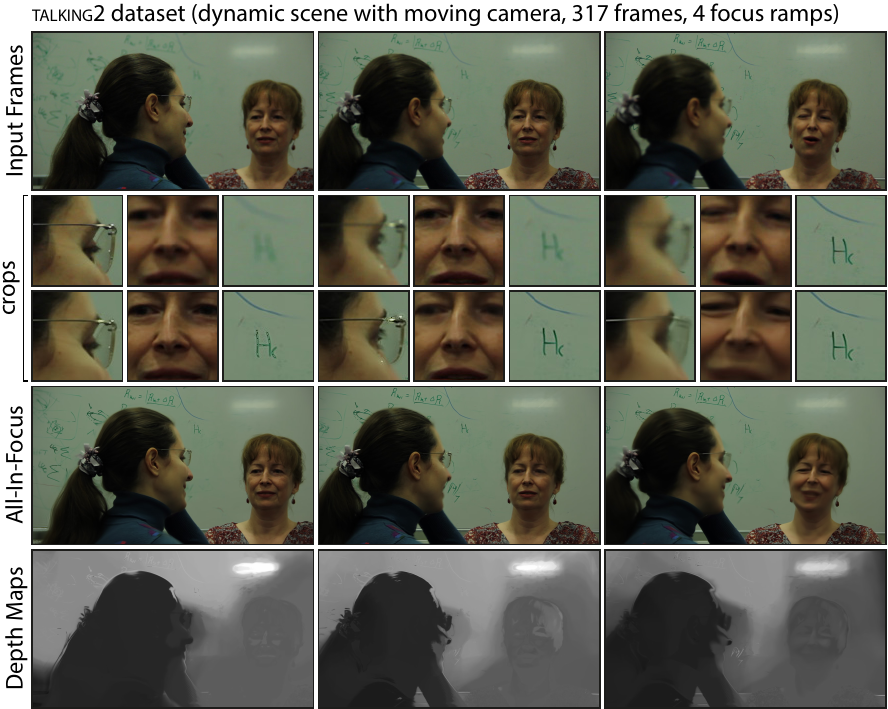}
\caption{
	Additional RGB-D video results.
	We show reconstructed all-in-focus images and depth maps for three focus sweep videos with various combinations of scene and camera motion.
	The image crops (top: input frame cropped, bottom: all-in-focus images cropped) focus on regions at the near, middle and far end (from left to right) of the scene's depth range.
	Note that each input frame is in focus in only one of the three crops, while our all-in-focus images are in focus everywhere.
	Please zoom in to see more details.
}
\label{fig:results-supp}
\end{figure}

\section{Evaluation of Focus Distance Refinement}
\label{sec:focus_refinement_evaluation}

\begin{figure}[t]
\centering
\includegraphics{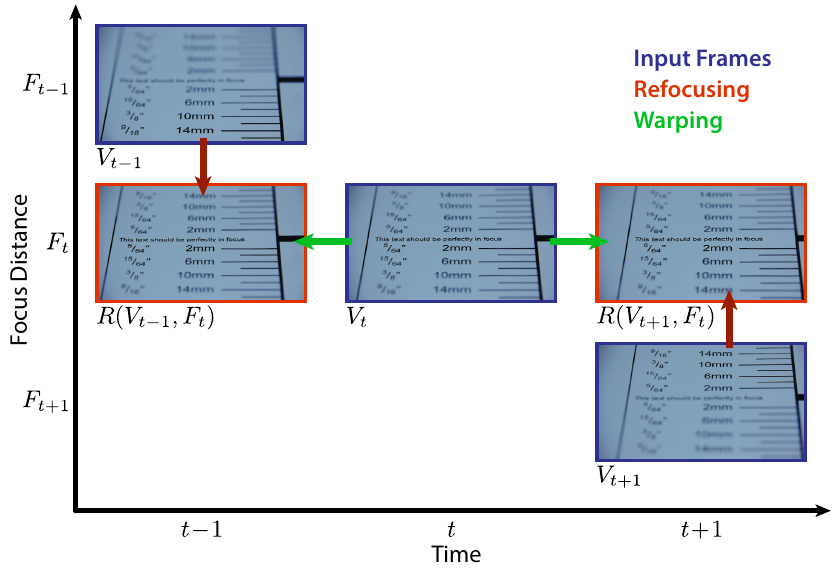}
\caption{\label{fig:focusOptimization}%
	We refine focus distances by refocusing input frames to each frame $t$, and minimizing the difference to the frame $V_t$ warped to each of the refocused input images (\protect\ifIncludeSuppMat{\cref{eqn:focusDistanceRefinement}}{\cref*{main-eqn:focusDistanceRefinement} in main paper}).
}\vspace{-1em}
\end{figure}

Here, we investigate the contribution of the focus distance refinement (\cref{fig:focusOptimization} and \ifIncludeSuppMat{\cref{sec:focus}}{\cref*{main-sec:focus} in the main paper}) to estimating better all-in-focus images and recovering from inaccurate initial focus distances.
For this, we process the synthetically refocused `alley\_1' dataset from MPI-Sintel~\cite{ButleWSB2012} with initial focus distances perturbed by varying degrees of additive Gaussian noise (but without imaging noise), with and without our focus distance refinement, and compare the all-in-focus images and estimated focus distances to the ground truth.
\Cref{fig:focusEvaluation} shows that our focus distance refinement consistently reduces the errors in estimated focus distances.
This in turn leads to better refocusing results for our defocus-preserving alignment (\ifIncludeSuppMat{\cref{sec:alignment}}{\cref*{main-sec:alignment} in the main paper}), which produces cleaner all-in-focus images and improves the overall performance of our approach.

\section{Applications of Video Depth-From-Defocus}
\label{sec:applications}

\inlineheading{Video Refocusing}
Given the estimated all-in-focus images and depth maps, we can now freely refocus the original input video according to the user's wishes by simply rendering the appropriate defocus blur in a post-process.
For this, we use the same thin-lens defocus model as in \ifIncludeSuppMat{\cref{sec:defocusmodel}}{\cref*{main-sec:defocusmodel} of the main paper}, and blur each pixel's neighborhood with the blur kernel $K(D(\mathbf{x}), F)$ corresponding to its depth $D(\mathbf{x})$ and the focus distance $F$ of the virtual lens \cite{RigueTI2003}.
This approach provides complete freedom, as the camera's aperture, focal length and focus distance can be changed independently and arbitrarily.
The user can for example change the aperture, while keeping the original focus settings, to reduce or magnify the defocus blur (see \cref{fig:refocusing}), similar to \citet{BaeD2007}, but for videos.
The focus can also be fixed on an object of interest or follow it through the video using a `focus pull', or the focus can be interactively controlled by the user using a `focus-follow' function that keeps the region under the user's mouse pointer in focus.
The reconstructed focus settings can also be smoothed to correct auto-focus failures and produce a more professional-looking result.

\begin{figure}[tp]
	\centering
	\includegraphics[width=\linewidth]{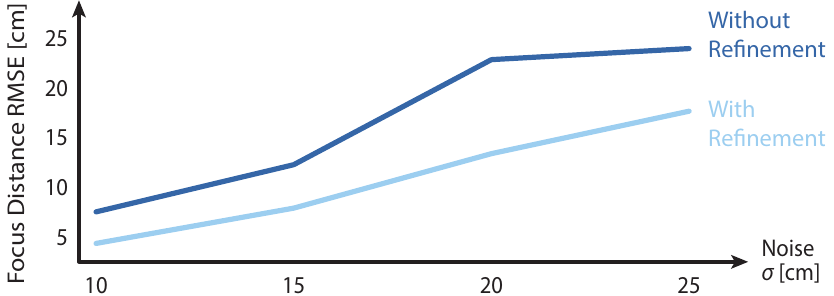}\\[0.75em]
	\includegraphics[width=\linewidth]{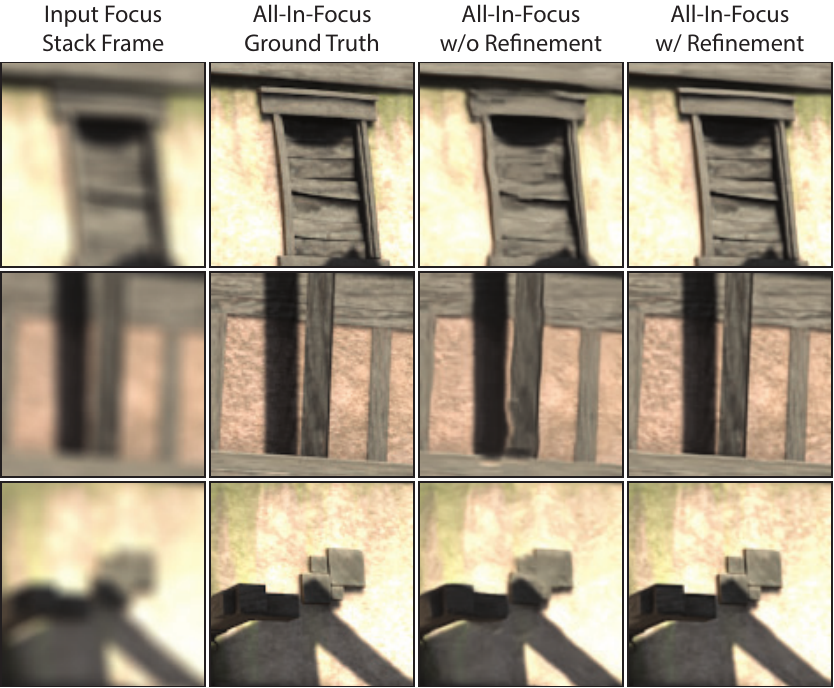}
	\caption{
		Focus distance refinement improves the focus estimates and all-in-focus images when the initial focus distances are inaccurate or noisy.
		\textbf{Top:} Plot of noise level versus RMSE of focus distances compared to the ground truth; note that refinement consistently reduces the error.
		\textbf{Bottom:} Crops of a single frame for noise level $\sigma \!=\! \text{10\,cm}$.
		Without refinement, the all-in-focus images are distorted and lack details; with refinement, the image is close to the ground truth.
	}\vspace{-1em}
	\label{fig:focusEvaluation}
\end{figure}

\begin{figure*}[t]
	\includegraphics[width=\textwidth]{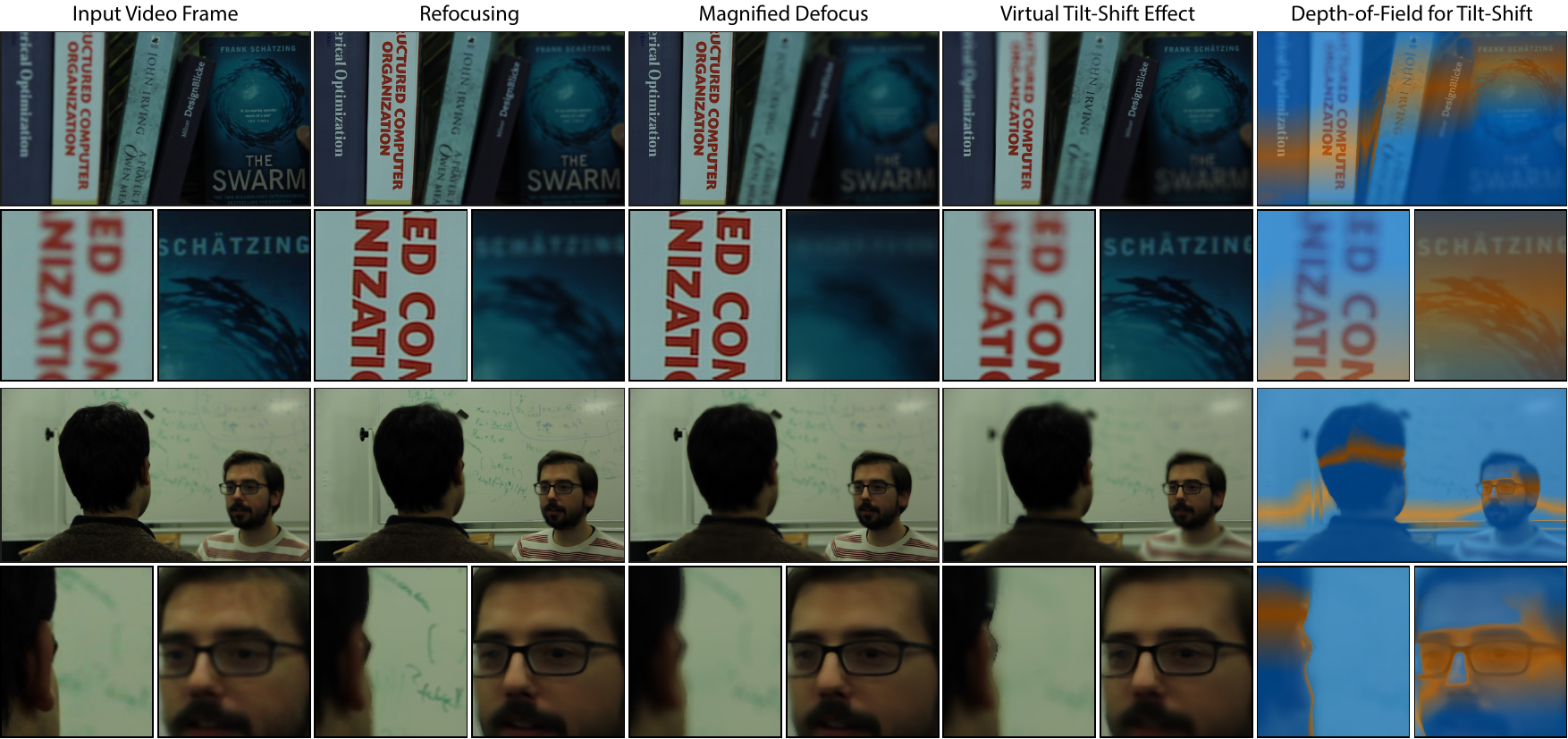}
	\caption{
		\textbf{Video refocusing results.}
		We first synthetically refocus the input video, then increase the defocus blur by increasing the aperture (smaller \textit{f}-number), and finally apply a virtual tilt-shift effect, which results in a slanted focus plane.
		Please see our video for full results.
	}\vspace{-1em}
	\label{fig:refocusing}
\end{figure*}

\inlineheading{Tilt-Shift Videography}
The \textit{tilt-shift effect} is created by tilting the camera's lens relative to its image plane which results in a slanted focus plane with a wedge-shaped depth of field that produces the iconic miniature look \citep{HeldCOB2010}.
(The purpose of lens \textit{shift} is to correct for perspective distortions like converging parallel lines; however, it does not affect the focus plane or depth of field.)
While the lens in most \textit{view cameras} can be tilted and shifted freely thanks to the flexible bellows between lens and film, most lenses in modern cameras are fixed to be parallel to the image sensor, which prevents this effect.
There are some special-purpose tilt-shift lenses for modern cameras, e.g. from Canon, Nikon or Lensbaby, which can be expensive, but the tilt-shift look is baked into the recorded footage and cannot be modified after capture.
We show virtual tilt-shift videography in \cref{fig:refocusing} and our video by refocusing with a tilted virtual lens \citep{Merkl2010}.
This provides ultimate flexibility as the desired look can be modified and tweaked interactively.

\inlineheading{Dolly Zoom}
Depth maps also enable other applications such as limited novel-view synthesis.
When combined with the video refocusing presented earlier, this provides the two ingredients required for a dolly zoom (or `Hitchcock Zoom'): a camera on a virtual dolly that moves towards or away from the scene, and a carefully controlled virtual camera zoom that keeps an object of interest at constant size (see supplemental video).
Assuming thin-lens optics, this is achieved by varying the focal length $f$ and object-to-lens distance $u$ such that the magnification $M \!=\! f / ( u \!-\! f)$ remains constant for the selected object.

}{}

{\small
	\bibliographystyle{ieeenat}
	\bibliography{VideoDepthFromDefocus}
}

\end{document}